# Metacognitive Agents for Ethical Decision Support: Conceptual Model and Research Roadmap

Catriona M. Kennedy
School of Computer Science
University of Birmingham, UK
catm.kennedy@gmail.com

**Abstract**. An ethical value-action gap exists when there is a discrepancy between intentions and actions. For example, people who support environmental sustainability often use cars and short-haul flights because of convenience and time-pressure. This discrepancy may be caused by social and structural obstacles as well as cognitive biases. Current technology can make this worse. For example, social media tends to enhance emotions such as anger or fear, which can result in polarisation and impulsive decisions.

Computational models of cognition and affect can provide insights into the value-action gap and how it can be reduced. Such models include dual process architectures, emotion models and behaviour change theories. In particular, metacognition ("thinking about thinking") plays an important role in many of these models as a mechanism for self-regulation and for reasoning about mental attitudes.

This paper outlines a roadmap for translating cognitive-affective models into assistant agents to help make value-aligned decisions. Key principles include "agile" rapid-prototyping using agent-based simulation, and the combination of descriptive and normative models into a single agent architecture.

## 1 Introduction

The value-action gap (or "intention-behaviour gap") happens when people take actions that are not consistent with their ethical values. The term is most often used in the context of ethical consumption and sustainability (e.g. Kollmuss and Agyeman, 2002). For example, many people agree that flying is bad for the environment, but take frequent short haul flights because of cost or convenience.

The gap can also be applied to values and actions more generally, and does not just apply to individual actions. For example, the values of an organisation might state that all employees should be respected and listened to, regardless of their power or status. However in practice there may be a failure to act in cases of bullying or harassment. Similarly, an organisation may support diversity, but make unfair hiring decisions. Current technology tends to widen this gap. For example, the algorithms of social media tend to enhance emotions such as anger or fear, and have a negative effect on reasoning and critical thinking.

An important goal of our research is to counteract the undermining of human independent reasoning by non-transparent algorithms and to support thoughtful ethical decision-making. The research is also relevant to the design of ethical AI systems in general. It is particularly important that AI systems should not learn human values from general human actions and text (such as social media), since these often contradict ethical values (and have their own value-action gaps). See for example, https://futurism.com/delphi-ai-ethics-racist.

Instead, we propose a decision support system that can help humans to reflect and debate about ethical questions. This in turn can provide a regulated environment in which machine learning about ethics can take place. Using the definitions in (Tolmeijer et al., 2020), the proposed decision support system fits into the category of an *explicit ethical agent* which is initially *top-down* and *deontological* (rule-based). In later stages of development, the system may incorporate some bottom-up elements. The top-down ethical requirements are elicited by a participatory value-sensitive design process (Dignum, 2019; Friedman and Hendry, 2019). These requirements are encoded in a knowledge representation that enables agent reasoning and explanation. For example, it can take the form of compliance-checking, with explanations for any violations. The agent's knowledge-base acts as a foundation for further learning about ethics.

We argue that cognitive-affective models can offer valuable insights for the design of decision support agents and that agent-based simulation can provide an environment for rapid-prototyping. A key point is the combination of descriptive and normative models, where descriptive models simulate actual features of natural systems and normative models simulate the required functionality of an artificial system to be coupled with the natural system.

The paper is organised as follows: section 2 covers related work; section 3 outlines desirable capabilities for an assistant agent; section 4 summarises key concepts in cognitive-affective models, including metacognition; section 5 outlines a roadmap for developing a decision support agent from cognitive-affective models and identifies key research challenges; section 6 presents a summary and conclusions.

## 2 Related Work

Symbolic cognitive-affective architectures are a foundation for our work. For example, H-CogAff (Sloman et al., 2005) emphasises the interaction between different layers of human cognition, namely the reactive, deliberative, and meta-management layers respectively. (Hudlicka, 2019) introduces a methodology for modelling the interactions between cognition and emotion in symbolic architectures. In particular, Hudlicka's contrast between "research" and "applied" models is similar to our distinction between descriptive and normative models respectively.

Symbolic decision support systems (Huang et al., 1993) provide a qualitative, argumentation-based decision process, which is closer to how humans actually make decisions in contrast to a purely quantitative optimisation approach. The CREDO stack (Fox et al., 2013; Fox, 2015) is a further development of this theory and provides valuable insights for our work.

Behaviour change assistance systems have similar goals to our research, since their aim is often to close a value-action gap. In particular, those systems that use reasoning or argumentation are relevant. An example of such an approach is STAR-C (Lindgren et al., 2020), which is a platform for intelligent coaching using goal-setting and activity modelling. In contrast to typical behaviour change assistants, however, personalisation is not the primary aim of our research. Instead, collaborative learning and shared experiences are important goals (although certain kinds of personalisation or tailoring are not ruled out).

## 3 Desirable Capabilities

In order to define capabilities, it is first important to list some scenarios where decision support might be used. Examples are as follows:

1) support for individual behaviour change, such as journey planning with sustainable transport.

2) helping individuals or groups to make ethical decisions on public policy options.
3) support for organisational decisions (e.g. hiring, employee monitoring, decisions affecting vulnerable groups, such as patients receiving mental health care, or claimants of social security benefits)

Reasons for value-action gaps in those scenarios can be identified. Support for public policy can be affected by biases or misinformation (for example, about under-served groups). Members of organisations can also lose sight of values due to work or business-related pressures.

**3.1 High level architecture**

Figure 1 gives a high-level view of desired agent capabilities. Some concepts from the CREDO agent specification (Fox et al., 2013; Fox, 2015) are used here as a framework. CREDO is based on the "Beliefs, Desires and Intentions" (BDI) model (Georgeff et al., 1999) but includes additional concepts such as goals and decision options. A decision option can be a programme of action or a belief decision (such as a medical diagnosis). Each option is associated with a list of arguments for and against it. Arguments are statements which, if true, give reasons why the option should be considered (or not). The decision process is a method of weighing up arguments for/against each option (e.g. arguments may have different positive or negative weights which are summed up, or a single argument may confirm or exclude an option without further evaluation).

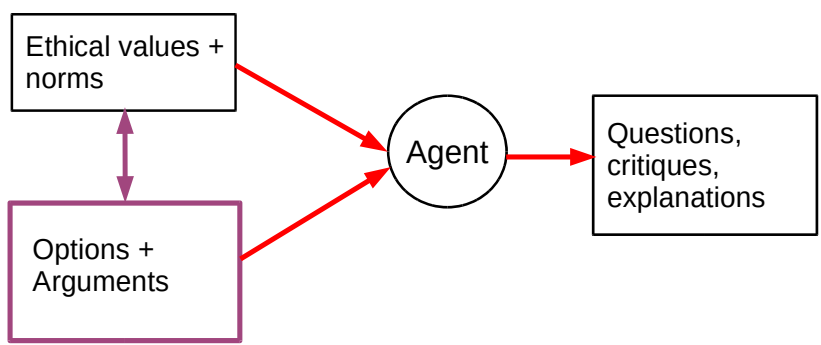

*Figure 1: High level architecture of a decision support agent*

The inputs to the agent are "ethical norms and values" and "options and arguments". The two-way arrow between them indicates consistency-checking. There may be situations where options and arguments are generated by the agent, but this is omitted from the diagram for simplicity. The agent outputs include "questions, critiques and explanations" and are explained in the next section ("Decision support functionality").

**Ethical values and norms:** Values can be defined as states or actions that are considered desirable in society (e.g. fairness, honesty). They are generic and apply to a wide range of scenarios (Schwartz 2011). Values may or may not be ethical. For example, fairness and honesty are ethical values, while others such as health or self-development are not. We are focusing here on ethical values that involve relations with others – such as other humans or animals, or "the common good" (e.g. human rights, conservation of the planet, future generations).

We use the definition of **norms** in the "Design for Values" framework in (Dignum, 2019, pages 63-67). In this framework, norms are specific interpretations of values. For example, fairness (a value) can have different interpretations, such as "equal access to resources" or "equal opportunities" .

Ethical values and norms are held in a knowledge-base. We are assuming that values may sometimes be included to provide a context for a norm (or a reason for it). Computational representation of some values can be approximated by specifying requirements for inter-agent relations (for example, honesty may be specified using rules). Deontological ethical theory is assumed here, with top-down specification in the early stages of development.

**Options and arguments:** Options and arguments may be specified in advance as part of a knowledge engineering process (as with CREDO for medical knowledge) or they may be agent-generated. A hybrid system combining top-down knowledge with bottom-up information extraction and learning is the aim here.

Depending on the scenario, an option can be an individual action or government policy (in the case of policy-decision-making). For example, an action can be a travel route involving walking or cycling, while a policy can be new regulation to restrict Covid-19 transmission. Action options can be evaluated as positive or negative according to values. They may be potential actions (possible future) or actual actions (current or past). An example of an actual action can be a current practice (such as unsustainable travel or food waste) which must be changed.

Arguments may be related to facts or values. Example argument patterns are: "It is good that we do X because it supports sustainability" (related to a value or norm) or "Option X involves Y, which gets us closer to the goal" (fact-related). We are using the term "argument" in an informal sense without making a commitment to argumentation as a paradigm.

### 3.2 Decision support functionalities

Figure 1 includes "questions, critiques and explanations" as part of the agent's output. They are best explained by defining some key functionalities as user interactions:

(1) **Option recommendation**: the agent presents the user with a list of options, and makes recommendations from the list, giving explanations as to why they are consistent with values. Explanations may be in the form of arguments in favour of a recommended option (or arguments against those that are not recommended). Options and arguments may be specified in advance using knowledge engineering (as in CREDO described above) or they may be discovered by the agent (depending on the complexity of the scenario).

(2) **Option critique**: the user is invited to present their proposed option(s), along with any supporting arguments. The agent checks if each option is consistent with ethical norms, giving explanations. This may include questions to the user, which they can then investigate. For example if the option is a product, does it include disposable plastic? Did the product involve exploitation of labour?

In both cases, the questions and explanations can help to achieve the following:
   (a) Draw attention to important issues that are being ignored or are invisible.
   (b) Explain why an option or argument may be the result of a bias or an affective state (explained further below).

A longer term goal is to refine representations of ethical values/norms based on mutual human/agent learning. This needs agreement on what kinds of refinements can be made to the representation of values or norms, and under what conditions these can take place.

### 3.3 Levels of autonomy

Interactive decision support can be part of a larger development process, with different goals. For example, the goal may be **fully automated decision-making**, but the agent is initially deployed in a semi-automated mode so that its recommendations can be corrected by a human. This semi-automated mode may be a "training phase" where the agent learns from mistakes. The final agent may be either a non-interactive decision system (such as credit-scoring or fraud detection) or an interactive system or chatbot that can give expert advice to non-experts (for example a medical advice system). In both cases, the agent's decisions have to be reliable, since they are not corrected.

Alternatively, the goal may be **human-agent collaboration** and mutual learning. In this case, the human and agent can correct each other as part of operational use (not a "training" phase).

### 4. Cognitive-affective models - background

Cognitive-affective models represent interactions between cognition and affect. We are focusing on the following key concepts: (1) *dual process architectures*, (2) *biases and affective states*, and (3) *metacognition* (both human and computational). These are elaborated below.

### 4.1 Dual-process architectures

In psychology, these architectures have two layers and are often called "system 1" and "system 2" (Kahneman, 2011). System 1 is fast but can be biased, and is associated with emotion and intuition; system 2 is slower and enables deliberate reasoning and reflection. System 1 can have a large influence on decision-making because system 2 requires effort and uses up energy, making it less able to challenge the biases of system 1.

Some behaviour change theories (BCTs) propose a similar multi-layer cognitive and motivational architecture. For example, CEOS (Borland, 2017) has many similarities to Kahneman's dual process model.

Computational cognitive architectures often make a similar distinction between "system 1 and "system 2", except that they are usually called *reactive* and *deliberative* processing respectively. An example is H-CogAff (Sloman et al. 2005). A key point is that reactive processing responds to information that is already available, while deliberation can reason about hidden information and generate hypothetical scenarios.

### 4.2 Biases and affective states

Many biases result from the effects of system 1. Some are particularly relevant for the value-action gap. (Kahneman, 2011) points out a number of biases that fall into the category of WYSIATI ("What You See Is All There Is"). In particular, a*vailability bias* is a tendency to be influenced by information that is visible, and to ignore hidden information (for example, in memory, in a news report or in a product description). This bias is related to *base-rate fallacy* and *conjunction fallacy* (both are a tendency to focus on specific examples while neglecting general information in the background). Similarly *framing* happens when an option is presented in a particular context (or "frame"). The framing of an option may emphasise positive or negative information, causing a decision-maker to select or avoid the option respectively. In this case, the frame is the "visible" information, while the invisible information (such as how the option could be presented in a neutral or negative way) has less influence. Availability-related biases can cause a value-action gap because we forget about important facts or values when other pressures or requirements are more visible.

Egocentric bias (Ross and Sicoly, 1979) is any kind of bias where we make decisions according to our own experience and world view, but fail to consider the experience or perspective of others. There are many related biases. An example is actor-observer asymmetry (Jones and Nisbett, 1972), where an observer neglects external circumstances as an explanation for the actions of others (and assumes they are due to character), but would emphasise external circumstances to explain their own actions in the same situation. Also related to egocentric biases are in-group biases, where we are more understanding about members of our own group. Perspective-taking requires cognitive effort (system 2) and people are more vulnerable to this bias in high pressure environments (Gilbert, 1989).

In addition to cognitive biases, emotions such as fear or anger can also affect decisions. A detailed discussion of emotion models would go beyond the scope of this paper, but two important concepts are relevant here: *appraisal theory* (Scherer et al., 2001) and *action preparation* (Frijda et al., 2014). Appraisal is a cognitive process which generates emotion in response to a significant event (e.g. an opportunity or threat). This process combines simple evaluation (positive or negative) with more complex reasoning, for example about the consequences of events or actions (Ortony et al., 1988). Action preparation happens as part of an emotional response and may include mental preparation, such as focusing attention on perceived threats, which can in turn cause biases (e.g. Hudlicka, 2007).

### 4.3 Metacognition

In psychology, metacognition plays an important role in therapy and self-regulation. For example, a person may become aware that their reasoning is affected by anger and they use a strategy to manage their anger using reappraisal (Gross and Thompson, 2007). The self-regulation model S-REF (Wells & Mathews, 1996) has been applied to psychotherapy for mental health disorders (Wells, 2011) and addictions (Spada et al., 2015). Metacognition is also important in bias correction (Wegener et al. 2012) and in social judgment (Petty et al. 2007), which makes it also relevant for ethical reasoning.

In computational agents, metacognition involves two components (Cox and Raja, 2011):
- *Object-level*: monitoring and control of events and processes in the environment
- *Meta-level*: monitoring and control of object-level information processing: e.g. error detection and correction, recognising knowledge gaps, setting new learning goals.

The nature of the information processing under scrutiny depends on the cognitive architecture. For example, if the architecture is dual process, it will include reactive and deliberative processing, as well as any affective processes being modelled.

For applied AI systems (agents/robots), components of metacognition can be included to make the system more robust or autonomous. Reasoning and explanation are important characteristics of metacognition in AI (Cox and Raja, 2011). Distributed and mutual metacognition is possible among multiple agents or within a single agent as part of a distributed cognitive system (Kennedy, 2011).

### 5 Methods and Roadmap

Cognitive-affective architectures can be applied in two ways: (a) to understand and model the problems that need to be solved (such as biases and the effects of emotion on cognition) and (b) to specify how to solve these problems – what kind of cognition is required instead? For (b) we are focusing on metacognition, although certain kinds of emotion models (such as empathy) may also have a role. Methods for both kinds of application are elaborated below, along with a high-level research roadmap.

**5.1 Role of agent-based simulation**

Agent-based simulation is useful for rapid-prototyping of designs and for early recognition of problems, as in agile software development. An agent-based simulation contains the following components:
- Scenario: usually a simulated environment with events, but may include real data
- Informal ontology: What objects and agents exist? What are the possible actions and events?
- Initial data: What is the starting state of the world and agents?
- Models: How do agents make decisions and act? How does the world change?

Simulation toolkits need to support cognitively rich agent modelling, so that experiments can be conducted with a variety of agent architectures (such as different emotion models, affective biases, or types of metacognition). The required tools may be part of a single simulation toolkit or they may be provided by a more specialist cognitive agent development toolkit which can be connected to simulated scenarios (or to the real world). Current examples of cognitive agent tools include GOAL (Hindriks, 2021). Although no longer supported, COGENT (Cooper & Fox, 1998) and SimAgent (Sloman & Poli, 1995) use important principles which can be applied in the development of new frameworks. In particular, they do not implement a single cognitive theory, but allow uncommitted exploration with multiple paradigms.

For most of our scenarios, simulation of mental states and information sources are of primary importance. Physical and spatial simulation are secondary (although they may be important for some public policy scenarios, such as house building or transport planning). In the early stages, it may be useful to test experimental cognitive agents in simple spatial environments for the purposes of easy visualisation (e.g. robots helping each other in a joint physical task), even if the final goal is co-operation in an information processing task, such as joint decision-making.

Participatory agent-based simulation is a key foundation for developing scenarios and agents (e.g. Taillandier et al., 2019; Ramanath and Gilbert, 2004). Recent work on Covid-19 simulations is particularly relevant for public policy scenarios (Dignum, 2021).

**5.2 Broad-and-shallow agents**

Early experimentation takes place on level 1 of Marr's cognitive architecture levels (Marr, 1982). Level 1 defines components and their interactions without specifying algorithms or representations (in other words, it specifies "what" is to be done, but not "how"). In practice, this means that early implementations of the agent architecture will have minimal components, which are populated with provisional and simplified algorithms. This is the "broad and shallow" approach (Bates et al., 1991), which is also the "agile" approach (early integration and ease of making changes). "Broad" means that the simulated agent has all the capabilities of cognition (perception, decision, action) as well as the essential components of the model under investigation. For example, if these components include reactive and deliberative layers, they need to be included in the simulation. However, these components are minimally implemented (in other words "shallow"). A shallow agent can be progressively "deepened" by replacing its algorithms with improved (or different) approaches as required (in which case, Marr level 2, and later 3, become the focus).

In a similar way, experimentation starts with simplified scenarios, which may be incrementally increased in complexity or replaced with radically different versions (e.g. changing from simple spatial worlds to virtual organisations or information only worlds).

## 5.3 Example: ethical workplace

For illustration, we will use an example scenario where managers in an organisation make decisions about employees (fits in with scenario 3 above). Depending on the business environment, managers may decide to change the workload of employees or to make redundancies.

We assume that managers wish to make ethical decisions and are respectful of employee well-being. However, in practice, this may not happen because of limited cognitive resources in a high-pressure environment, meaning that "system 1" will have most influence on decisions. Since system 1 only responds to information that is directly available, the availability bias will cause the manager to neglect the perspectives of employees who are not "visible" (thus causing a value-action gap).

In a simulation, managers and employees can be represented as agents with a dual-process architecture (reactive and deliberative layers) which are subject to biases. Each staff role can be represented as an agent type with its own perspective, which can be defined as its interpretation of the world, its goals, needs, and available actions. An "interpretation" can be understood as the agent's system of concepts (like an ontology but not necessarily a formal ontology). To simulate the availability bias, manager decision processes can prioritise information that is immediately available to them (e.g. raised in meetings with other managers) over information that is invisible, and cause the ethical values themselves to be temporarily forgotten.

A simple way to introduce bias or emotion is to enable the reactive layer to be immediately activated by visible information that might be a threat or opportunity (e.g. using an appraisal model). This results in impulsive decisions being made before the deliberative layer has time to reason about options and arguments.

This scenario does not include all the complexities and conflicting pressures of a real-life workplace. However, other features and biases may be incrementally added, including simulation of emotions and affective biases.

At this stage, we can define the following research question:

**RQ1: How to model cognitive-affective biases in a symbolic descriptive way that is easy to understand for stakeholders and fits into a rapid-prototyping framework (i.e. easy to change and experiment with).**

There are many possible sub-questions associated with RQ1. For example, should the models of bias be adapted to specific situations or agent perspectives? These questions can be defined more precisely in later iterations of development.

## 5.4 Descriptive vs normative models

The simulation so far is descriptive. The next stage is to explore designs for a potential assistant agent. For this purpose, the distinction between descriptive and normative models is important:

- **Descriptive**: what features do natural systems have, including problematic features, such as biases?
- **Normative**: what features do we want (or what ought to happen)? These are software requirements. For ethical agents, "normative" includes alignment with ethical norms.

There are two important considerations: *first*, who determines the requirements that a "normative" system satisfies? For ethical software systems, this needs to be a democratic and participatory process, delivering norms as an output. Examples are "Value Sensitive Design" (Friedman and Hendry, 2019) and the "Design for Values" process of (Dignum, 2019).

*Secondly*, "normative" is not necessarily the same as "optimal". Some biases and shortcuts in reasoning may be helpful, particularly when fast decisions are required with limited information, and when small errors have no significant consequences. This is the idea of "satisficing" (Simon, 1956).

**5.4.1 Role of metacognition**

Within the context of metacognition, we are defining a "normative" system as one which satisfies requirements in realistic scenarios where biases and other problems exist. In order words, we need a corrective process which is embedded inside the biased, error-prone system represented by the descriptive model. Before this bias correction can take place, a normative (desired) decision process needs to be specified, according to ethical values and norms. This can be labelled M0. For example, it might specify the requirement to reason about a decision's possible consequences for vulnerable groups (in the Public Policy scenario).

These considerations lead to three classes of agent models:
M0: normative: simulate the required decision process (no biases).
M1: descriptive: simulate biases, with absent or ineffective metacognition.
M2: normative (added to descriptive): simulate the same biases, but with effective metacognition.

Model type M2 is an autonomous agent that is derived from M1 but is effective in correcting its own biases to the extent that requirements are satisfied. (This is not the same as "perfect" or "optimal"). Effective metacognition means that it monitors its actual reasoning processes (M1), compares them with the required processes M0, and makes adjustments as necessary. In this case M1 is the "object-level" of M2 (as defined in 4.3). The corrected process does not have to follow M0 exactly, so long as it satisfies the functionality of M0 on a high level (Marr level 1).

Additionally, metacognition should enable the agent to *explain why* its decision process (M1) was problematic (using introspection) and what deliberation process it activated instead (using metacognitive control). For example, in a rule-based system, the agent may have detected its preparation to act on a decision that was made impulsively, where "impulsivity" can mean triggering a reactive rule without considering relevant, but hidden information.

This leads to the following research question:

**RQ2: How to add metacognition to a biased agent so that the agent can reason about its biases and use metacognitive control to counteract them.**

Additional sub-questions of RQ2 involve the precise specification of M0. This includes the representation of ethical norms and the possibility of learning some details of how to satisfy them. For example, this may include priority allocation to options under consideration.

**5.5 Autonomous vs assistant agents**

It is useful to define an autonomous agent initially to allow us to focus on the required processing in the cognitive architecture without having to consider how to communicate this to other agents or how to design human-agent interaction. The next question is whether an autonomous agent can be translated into an assistant agent by applying its reasoning about itself to other agents. For this purpose, we can define an additional model type M3, which is derived from M2:

M3: same as M2, but uses its metacognition to give advice to another agent of type M1.

In the most simplified version, the assistant agent shares the same "experience" of the environment as M1. Therefore M3 can challenge M1's decisions using shared mental concepts that are not specific to a particular agent's experience. Since the environment is shared, both agents can make a decision on the same event, except that M1 proposes a biased option while M3 gives advice on how to make a correct decision.

M3's advice can include an explanation of how it corrected its own reasoning. If we assume that M1 has ineffective (or inactive) metacognition, an explanation can direct M1's metacognition to become aware of flaws in its reasoning. For example, M1 could propose an option A, which goes against an ethical norm. The challenge from M3 can then have the following pattern: "I first thought option A was the best choice, but I realised this decision was impulsive (or was affected by availability bias) and I had not considered other relevant information which makes it clear that the option is unethical. I propose option B instead." The agent could then state its arguments for/against option A and those for/against option B and showing how it reached its decision using deliberation. In particular, if one argument shows that an option violates an ethical norm, the option should be excluded immediately (without the need for further weighing up of other arguments). In the same way, specific arguments for or against an option can also be challenged. For instance, if ethical norms are never mentioned in any arguments used by M1, this absence may be pointed out by M3.

These kinds of advice and explanation cover functions (1) and (2) of the desirable capabilities above (namely: recommending options, challenging a proposed option or argument). In addition, the explanations provide features (a) and (b), namely, drawing attention to hidden information, and raising awareness of biases or affective states that are likely to affect decisions.

This leads to the following research question:

**RQ3: to what extent can an autonomous metacognitive agent be translated into an assistant agent which gives advice to another agent by explaining its own reasoning?**

The above assumes that M3 has the same environment and experiences as M1. Additional sub-questions of RQ3 involve different agent experiences and the capability of agents to reason about other perspectives.

**5.6 Human-in-the-loop simulation**

The next question is how to translate an agent-to-agent interaction into a human-agent interaction. This would mean that a human user plays the role of the decision agent M1 while the adviser agent becomes the prototype decision support system. Existing work in participatory and role-playing simulation is important here (e.g. Guyot and Honiden, 2006).

We can define a new model type M4 which is derived from M3, but with the additional features necessary for human-agent interaction. M4 uses a representation of M1's reasoning, which might also act as a generic model of the human user (without personalisation). The goal is to use a generic representation of known biases that affect human decisions, beginning with availability bias. The research question can be defined as follows:

**RQ4: to what extent can an agent that advises a simulated agent be translated into an agent that can advise humans?**

The limits of the generic model M1 (used by M4) can be identified and fed back to stage 1 (descriptive agent modelling).

Additional sub-questions of RQ4 involve the measuring of effectiveness of an assistant agent: does it actually draw attention to invisible information, and does it really help people to be aware of biases? This is in accordance with decision support goals (a) and (b) in section 3.2. Mutual human-agent learning would be developed at this stage. For example, the assistant agent initially only has a generic model of a biased decision-process (only one kind of M1) and does not have variants of this model to fit specific situations or persons. (Although personalisation is not a goal in this research, some degree of adaptation or tailoring may be helpful).

**5.7 Research roadmap - summary**

The roadmap stages can be summarised as follows:

1. Simulate problems, such as biased decisions.
2. Simulate solutions by adding metacognition: determine desirable capabilities for correcting problems, the role of metacognition, and how to evaluate its effectiveness.
3. Translate from autonomous agent to assistant agent.
4. Introduce *human-agent interaction*. For example, humans could interact with assistant agents where the scenarios are still simulated. Results can be used as input for building a non-simulated decision support system.

The stages are shown in Table 1, along with model outputs, research questions and possible next stages. Since the process is iterative, the next stage may be a previous one. This is indicated by "[ ]". At each stage, additional research questions can be defined, but since we are using the "broad-and-shallow" methodology, the goal is to develop a minimal complete system as quickly as possible. This enables evaluations to be made early and the resulting feedback to be used when returning to earlier stages.

*Table 1: Roadmap for agent simulations*

| Stage | Model output | Model type | Research question(s) | Possible next stage(s) |
| --- | --- | --- | --- | --- |
| S1: Simulate problems – biases, emotions | M1 | Descriptive | RQ1 | 2, 3 or 4 |
| S2: Simulate solutions – autonomous agent | M0, M2 | Normative | RQ2 | 3 or 4 [1] |
| S3: Simulate solutions – assistant agent | M3 | Normative | RQ3 | 4 [2 or 1] |
| S4: Build human-agent interaction | M4 | Normative | RQ4 | n/a [3, 2 or 1] |

**6 Summary and Conclusion**

In this paper, we identified scenarios where value-action gaps exist, particularly in the context of ethical decision-making. These do not just include individual behaviour change scenarios (such as sustainable consumption) but also organisational and policy decisions. We then defined some desirable capabilities for an assistant agent (decision support system) to help close such value-action gaps.

Concepts from cognitive-affective architectures were identified as relevant for the design of assistant agents. These concepts include dual process architectures ("system 1 vs system 2"), cognitive biases, and metacognition. In particular, metacognition has relevance for ethical reasoning because of its role in self-regulation, social cognition, and in understanding the limits of one's own knowledge.

A high-level roadmap was then defined, with particular emphasis on agent-based simulations and the combination of descriptive and normative models in a single agent architecture. A major challenge is the level of resolution of models, particularly for emotion and bias simulation. Our approach is to start at a low level of resolution ("broad and shallow" agents) and iteratively add detail as necessary.